\pdfoutput=1
\documentclass[journal]{IEEEtran}
\usepackage{amsmath}
\usepackage{amsfonts}
\usepackage{threeparttable}
\usepackage[pdftex]{graphicx}
\usepackage{hyperref}
\usepackage{booktabs} % nicer looking tables
\hypersetup{
    colorlinks=true,
    linkcolor=blue,
    filecolor=blue,      
    urlcolor=blue,
    citecolor=blue,
}
\graphicspath{{images/}}
\DeclareGraphicsExtensions{.pdf,.jpg,.png,.svg}
\begin{document}
\title{Model-based graph reinforcement learning for inductive traffic signal control}
%\title{Flexibility and coordination in large multi-agent action spaces}
\author{François-Xavier Devailly, Denis Larocque, Laurent Charlin
\thanks{F-X.\ Devailly, D.\ Larocque, and L.\ Charlin are with the Department of Decision Sciences at HEC~Montr\'eal, Qu\'ebec, Canada. E-mail: francois-xavier.devailly@hec.ca, denis.larocque@hec.ca, laurent.charlin@hec.ca. This work was partially funded by the Natural Sciences and Engineering Research Council (NSERC), Fonds de Recherche du Qu\'ebec: Nature et Technologies (FRQNT), Samsung, and Fondation HEC Montr\'eal.}}%

\maketitle
\begin{abstract}
Most reinforcement learning methods for adaptive-traffic-signal-control require training from scratch to be applied on any new intersection or after any modification to the road network, traffic distribution, or behavioral constraints experienced during training. Considering 1) the massive amount of experience required to train such methods, and 2) that experience must be gathered by interacting in an exploratory fashion with real road-network-users, such a lack of transferability limits experimentation and applicability. Recent approaches enable learning policies that generalize for unseen road-network topologies and traffic distributions, partially tackling this challenge. However, the literature remains divided between the learning of cyclic (the evolution of connectivity at an intersection must respect a cycle) and acyclic (less constrained) policies, and these transferable methods 1) are only compatible with cyclic constraints and 2) do not enable coordination. We introduce a new model-based method, MuJAM, which, on top of enabling explicit coordination at scale for the first time, pushes generalization further by allowing a generalization to the controllers' constraints. In a zero-shot transfer setting involving both road networks and traffic settings never experienced during training, and in a larger transfer experiment involving the control of 3,971 traffic signal controllers in Manhattan, we show that MuJAM, using both cyclic and acyclic constraints, outperforms domain-specific baselines as well as another transferable approach.

\end{abstract}

%\section{Candidates}
%MuJAM: MuZero + Joint Action Modeling (same logic as AlphaFold)
%\newline
%GIM-JAM: Graph Inductive Model-based Joint Action Modeling
%\newline
%TRIM-JAM: TRansferable/Inductive Model for Joint-Action Modeling

\section{Introduction}
Adaptive-traffic-signal-control (ATSC) aims at minimizing traffic congestion which gives rise to a plethora of environmentally and socially harmful outcomes~\cite{schrank20122012,barth2008real,schrank20152015}. Reinforcement learning (RL), via trial-and-error, of ATSC policies, is popular to move beyond heuristic-based approaches which rely on both manual design and domain knowledge~\cite{hunt1981scoot, lowrie1990scats}. Multi-Agent RL (MARL) in particular, promises scalability of RL approaches by dividing control among intersections, but has mainly been restricted to training \emph{specialist agents} which can only be applied on the exact road-network-topology, traffic distribution, and under the same constraints as experienced in training (see \S~\ref{related_work}). This lack of transferability, combined with notorious data inefficiency of these methods (i.e. the need to gather a massive amount of experience in order to train RL policies)~\cite{wei2019survey} poses a challenge to real world applicability as the social acceptability of prolonged exploration via interaction with real road-network-users remains questionable. Furthermore, most MARL-ATSC approaches sacrifice coordination ability between agents (see \S~\ref{related_work}). 

\subsection{Contribution}
Model-based RL (MBRL), which explicitly models the dynamics of the environment tend to both outperform model-free RL (MFRL) in complex planning tasks and enable better sample efficiency~\cite{silver2017mastering,schrittwieser2020mastering}.
We introduce joint action modeling with MuZero~\cite{schrittwieser2020mastering}, MuJAM, a new family of MBRL approaches for RL-ATSC which models dynamics of the ATSC environment at the lane level using a latent space. 

%\textbf{Inherited Capabilities}
\subsubsection{Inherited Capabilities}
Most neural-network architectures do not enable dealing with changing number of inputs and state dimensionality. For this reason, in a traffic scenario where various types of entities such as cars enter, move inside of, and leave the network, these methods do not enable full granularity exploitation. Instead of using loop sensor information and resorting to manual and arbitrary aggregations, flexibility in the computational graph offered by graph convolutional networks (GCNs) enables the exploitation of available data at its finest level of granularity (e.g., at the vehicle level)~\cite{devailly2021ig}. These approaches leverage inductive capabilities of GCNs to enable transfer, with no additional training, to new intersections and traffic distributions. This in turn translates into massive scalability as demonstrated in related MFRL approaches~ \cite{devailly2021ig}.

On top of these inherited abilities, MuJAM outperforms its MFRL peers and offers, the following advantages: 
%\begin{itemize}
%\item \textbf{Acyclic-Inductive-Graph RL}: 
\subsubsection{Acyclic-Inductive-Graph RL}
RL-ATSC remains divided between the learning of cyclic and acyclic policies~\cite{wei2019survey}. This comes at a cost for urban-mobility-planning as experimentation with different constraints systematically requires retraining using a different approach. Even transferable approaches are limited in this regard. Only when the policy consists in switching to the next phase or remaining in the current phase (using cyclic constraints) does the action space remain binary and identical across intersections. For this reason, methods which aim to generalize over road-network architectures and traffic distributions are limited to the use of cyclic constraints~\cite{devailly2021ig}. MuJAM enables the use of transferable approaches, which were limited to cyclic constraints until now~(see \S~\ref{related_work}), with acyclic constraints. Policies under the latter, looser type of constraints, typically perform better.
%\item \textbf{Constraint-Agnostic ATSC}: 
\subsubsection{Constraint-Agnostic ATSC}
MuJAM pushes generalization ability further and a unique instantiation of this model can generalize over constraints and be used, for instance, with both cyclic or acyclic policies on any new intersection. This additional level of generalization and transferability is intended to ease urban-mobility-planning by limiting the costs of experimentation and application.
%\item \textbf{Coordination \emph{at scale}}: 
\subsubsection{Coordination \emph{at scale}}
It enables explicit joint action modeling \emph{at scale} for the first time ~(see \S~\ref{related_work})

%\item \textbf{Data efficiency}: 
\subsubsection{Data efficiency}
A) Improved Exploration: We adapt recent MBRL approaches to enable more efficient exploration which translates into better data efficiency as we gather more relevant signal with fewer interactions with the environment. B) Sample Efficiency: One of the main motivations behind MBRL is the ability to  facilitate learning and reach better sample efficiency by leveraging a model of the dynamics of the environment. MuJAM does enable greater sample efficiency compared to related approaches. 
%\end{itemize}

To test our claims, we introduce and evaluate various instantiations of MuJAM (and perform ablation studies) on zero-shot transfer settings involving road networks and non-stationary traffic distributions both never experienced during training, and show that our various instantiations outperform trained-transferable and domain-specific baselines

\section{Related Work}
\label{related_work}
%\textbf{Scalability}
\subsection{Scalability}
Even though some works propose to control a few intersections using a single agent~\cite{casas2017deep,prashanth2011reinforcement}, the explosion of both state and action space dimensionalities with the number of intersections limits the scalability of such approaches. MARL-ATSC aims at making RL scalable by decentralizing control~\cite{el2010agent,abdoos2011traffic,el2013multiagent,khamis2012enhanced,steingrover2005reinforcement,salkham2008collaborative,van2016coordinated,arel2010reinforcement,balaji2010urban,aslani2018traffic,pham2013learning,kuyer2008multiagent,zheng2019learning,xiong2019learning,nishi2018traffic,wei2019colight,xu2013study,khamis2012multi,aslani2017adaptive,chu2019multi,ivsa2006reinforcement,wiering2004simulation,wiering2000multi,zhang2019integrating,mannion2016experimental,wiering2004intelligent}. However, an increase of parameters leads to an explosion of costs (computational power, memory, and data requirements) with the number of intersections, which hinders scalability. Parameter sharing in MARL methods~\cite{devailly2021ig,zang2020metalight, zheng2019learning, wei2019colight,chen2020toward,wang2019stmarl} has enabled far greater scalability (up to a thousand controllers). Such methods usually rely on GCNs. Transfer learning is a promising way to reach even further scalability while drastically diminishing the training requirements. To limit the quantity of experience required to training under a new setting, ~\cite{zang2020metalight} uses meta-learning, and~\cite{devailly2021ig} achieves zero-shot transfer from small synthetic networks to massive real world networks using GCNs' inductive capabilities.
%\textbf{\underline{Generalization \& Transferability}}: \cite{zang2020metalight} uses meta-learning to limit the quantity of experience required to training under a new setting, and only \cite{devailly2021ig} enables immediate transfer (i.e. zero-shot transfer) to unseen settings.

%\textbf{Coordination}:
\subsection{Coordination} 
Most MARL approaches for ATSC methods offer either no means of coordination between intersections, or only the ability to communicate~\cite{el2010agent,nishi2018traffic,zhang2019integrating,wei2019colight,devailly2021ig,arel2010reinforcement}. Joint action modeling is challenging because of the non-tractability of exploding action spaces. Some approaches do tackle joint action modeling on small road networks~\cite{el2013multiagent, kuyer2008multiagent,van2016coordinated,xu2013study}.

%\textbf{Data Efficiency \& Model-Based RL}:
\subsection{Data Efficiency \& Model-Based RL} One of the main challenges to real-world applicability of RL-ATSC is the notorious data inneficiency of RL algorithms. This data inefficiency can be decomposed into 2 sub-challenges 1) Exploration: Guiding the collection of experiences to gather insightful observations/transitions in as few interactions with the environment as possible. 2) Sample Efficiency: Learning from the collected observations/transitions as efficiently as possible. Despite the fact that one of the main appeals of model-based RL (MBRL) is precisely to improve the latter, only a few works have explored such approaches for  ATSC~\cite{salkham2010soilse,khamis2012multi,wiering2000multi,wiering2004intelligent,steingrover2005reinforcement,khamis2012enhanced,wiering2004simulation,kuyer2008multiagent}, and this research avenue has not brought much attention since the early 2010s.

\section{Background}
\subsection{Model-Based Reinforcement Learning}
Sequential decision making can be framed as a Markov decision process (MDP). An MDP is defined by: $S$ a set of states, $A$, a set of actions,  $\mathcal{T}$, a transition function defining the dynamics of the system (the probabilities of transitioning from a state to another state given an action), and $R$, a reward function defining utility (i.e. performance w.r.t. the underlying task). Solving an MDP means finding  a policy, $\pi$, which maximizes the expected accumulation of future value (discounted by a temporal factor $\gamma$): $R_t=\sum_{k=0}^{\infty}{\gamma^kr_{t+k+1}}$ where $r_t$ represents the reward experienced at time step t. RL is a popular framework to solve MDPs via trial-and-error. When the learning of value and transition functions are confounded, RL is said to be model-free. Alternatively, MBRL approaches explicitly leverage transition dynamics to enable planning (i.e. simulating trajectories to select promising ones up-front). In MBRL, models of the dynamics which are perfectly known in advance can be provided to the RL method. This, combined with tree search (TS)\footnote{Tree search is an heuristic planning method simulating many roll-outs by randomly sampling the search space to identify and analyze the most promising actions.} planning, has enabled AlphaZero to reach superhuman performance in complex planning tasks such as Chess, Shogi, and Go~\cite{silver2016mastering,silver2017mastering}. However, in most real-world environments, dynamics are complex and unknown and must be learned. Learning transition functions (i.e. $s_{t+1} = \mathcal{T}(s_t,a_t)$) in high dimensional state spaces is expensive and challenging as the performance of MBRL approaches significantly lagged that of MFRL approaches in corresponding tasks~\cite{schrittwieser2020mastering}. Recent approaches aim to include implicit feature selection and dimensionality reduction by planning in a value-equivalent-latent-space, instead of planning in the original state space of the MDP~\cite{oh2017value,schrittwieser2020mastering}. In other words, a dynamic model learns, given a starting state, 1) to map the original state space to a latent (vectorized) space and 2) to simulate trajectories in this latent space under the only constraint that the prediction of value-related quantities based on the latent space must match those observed in the real state-space. Such approaches have 2 main advantages: 1) The dynamics model can ignore all dynamics unrelated to the task at hand (dynamics which do not influence value-related metrics), 2) computations are projected into a latent vectorized space instead of predicting (usually reconstructing) high dimensional states such as images.  With these advances, MuZero~\cite{schrittwieser2020mastering} matches the performance of MBRL algorithms in their favored domains (complex planning tasks) and outperforms MFRL algorithms in their favored domains (states involving high dimensional state spaces).

\subsection{Heterogeneous Graph Convolutional Networks (GCNs)}
Graph convolutional networks consist in stacking convolutional layers to recursively aggregate transformations of neighborhood-information in a graph~\cite{kipf2016semi}. As a message passing framework, GCNs enable the learning and exploitation of nonlinear patterns involving both structural and semantic (nodes and edges features) signal. Their extensions to heterogeneous graphs (with multiple types of nodes and edges), Heterogeneous GCNs or HGCNs~\cite{schlichtkrull2018modeling}, work by applying the following message-passing equation on every node, at every layer: 
\begin{equation}
    \begin{split}
        \begin{gathered}
            n_i^{\left(l\right)}=f\bigg(\sum_{e\epsilon E, j\epsilon N_e\left(i\right)}{W_{l_e}\cdot n_j^{\left(l-1\right)} \bigg)}
        \end{gathered}
    \end{split}
    \label{equation:R-GCN}
\end{equation}
where $n_i^l$ is the embedding at layer $l$ of node $i$, $f$ is a non-linear differentiable function, $N_e(i)$ is the immediate neighborhood of node $i$ in the graph of relation type $e$, $E$ is the set of relation types, and $W_{l_e}$ is the $l^{th}$ layer’s weight matrix for message propagation corresponding to a relation of type $e$. Parameters are typically learnt by backpropagation of an error obtained using nodes embeddings. For simplicity, we will refer to HGCNs as GCNs for the rest of the paper.

\subsection{ATSC Decision Process}

%\textbf{Decision Process}:
\subsubsection{Decision Process}
The decision process consists in the simultaneous and coordinated control of all TSCs in a given road network.

%\textbf{State}:
\subsubsection{State}
The state of a road network comprises 1) \emph{connectivity}: the status of all existing connections between lanes at intersections, and the status of their respective controllers' constraints, and 2) \emph{demand}: vehicles positions and speeds. 
State information is encoded using vector representations which qualify nodes and edges of the GCN. (see~\ref{sec:architecture}).

%\textbf{Action}:
\subsubsection{Action}
The joint action at a given time step, is a combination of local actions, each consisting in the selection of a legal\footnote{A legal phase is a phase which can be selected by a TSC at a given time step according to this TSC's constraints} phase among all legal phases. Once a phase is selected to be the next phase, one of two things happen: If the selected phase is the current one, then no switch is performed. On the other hand, if the selected phase is different from the current phase, the chosen phase becomes the target phase and the switch to this target phase begins. If an intermediary phase involving yellow lights is required (e.g. if some lights are green in the current phase and will be red in the destination phase), then this intermediary phase is enforced for a fixed duration before activating the destination phase.

%\textbf{Transition constraints}:
\subsubsection{Transition constraints}
A TSC can only select a phase different from the current one after a minimum duration has passed. All intermediary phases involving yellow lights last for a fixed duration. If a TSC is under cyclic constraints, then it can only select the current phase or the next phase in the cycle. If a TSC is under acyclic constraints, then it can select any legal phase. 

%\textbf{Reward}:
\subsubsection{Reward}
The reward is the negative sum of local queues lengths.\footnote{A vehicle is counted in a queue if it is stopped at a maximum of 50 meters of the intersection}

%\textbf{Step}:
\subsubsection{Step}
The duration of each step is one second. 

%\textbf{Episode}:
\subsubsection{Episode}
Episodes can either end after all trips are completed, or after a given amount of time.

\section{MuJAM}
\label{sec:MuJAM}

Our new family of approaches for RL-ATSC builds upon inductive-graph-reinforcement-learning (with GCNs) and models dynamics of the ATSC environment using a latent space to facilitate the combination of MBRL and GCNs. Modeling is decentralized at the level of lanes and performed jointly for all lanes in the road-network to enable full coordination (joint action modeling).

\subsection{Dynamics Model}
We distinguish two parts of environmental dynamics in the ATSC MDP.
\begin{enumerate}
    \item \emph{Connectivity dynamics} are simple and known in advance. If a given action is taken, connectivity (e.g. which connections between lanes at a given intersections will be opened) is influenced in a fully deterministic and predictable way. 
    \item \emph{Traffic dynamics} (i.e. the way vehicles behave under a given state of traffic and connectivity) are complex and unknown.
\end{enumerate}

\emph{Traffic dynamics} have to be learned. MBRL with a learned dynamics model usually involves learning to predict the next state conditioned on the current state and a given action. A challenge is that in the current setting, states are complex graphs with  varying number of nodes, connectivity, and features. All of these elements change from one time step to another (as vehicles move along the road network for instance). Learning to predict such a complex object in a learnable/differentiable way is non trivial. It becomes even more challenging if planning has to be performed in real-time which is required for real-world ATSC. For this reason, we use MuZero~\cite{schrittwieser2020mastering}, which enables planning in a latent vectorized space by only using value-related scalars as targets to learn the model. 

\emph{Connectivity dynamics}, on the other hand, are directly provided to the model as features in the GCN and are manually updated during model-based-planning. Note that the model is informed of the type of constraints used (via features in the GCN).

Our dynamics model is learned at the lane level. For planning, we start by obtaining an initial state representation for all lanes (a vector obtained using a GCN which aggregates information about structure, connectivity, traffic, etc.). Then, to model a transition given an action, we 1) manually update features related to connectivity dynamics in the GCN, 2) use the GCN to update representations (vectors) for all lanes, 3) predict rewards $r$ and long-term values estimates $v$ (i.e. predicted future cumulative reward) for all lanes based on their respective vectorized representations. The final value-related metrics ($r$,$v$) associated to a given transition for the entire road network is then simply obtained by summing all lane level estimates. 

\subsection{Coordinated priors for parsimonious planning}
\label{coordinated_priors}
In a coordinated setting, the action space rapidly becomes intractable with the number of intersections. The total number of actions at a given time step is $\prod_{i=1}^{n} |A_{i}|$ where $n$ is the number of intersections and  $|A_{i}|$ is the size of the action space for intersection $i$ according to its current state. Even for a small road network of 20 intersections, modeling all transitions for a single time step (i.e. performing an exhaustive search of the first layer of the TS) could require querying our GCN-based-dynamics-model more than a million times. Instead, we sample the action space parsimoniously. To do so, we adapt the TS sampling strategy proposed in Sampled MuZero~\cite{hubert2021learning} to a coordinated setting. Namely, we train a prior multilayer perceptron (MLP) $\phi$ to predict, at the intersection level, the local action which will be part of the best coordinated set (according to the results of the TS). 

\subsubsection{Local priors}
Logits are obtained using $\phi$ on phase nodes representations, and used to define multinomial distributions with parameters: $p_{i} =exp(l_{i})/\sum_{j \epsilon A} exp(l_{j})$, where $p_i$ is the probability of sampling action $i$, A is the set of actions (i.e. legal  phases) for the TSC at the current state, and $l_i$ is the logit corresponding to action $i$.
Each of these distributions can be used to sample local actions (i.e. actions for each TSC),
\subsubsection{Best coordinated set}
A TS is used to identify the best coordinated set of actions among all candidate sets. To sample a candidate set, a local action is sampled for all intersections using the local multinomial distributions. The best candidate set (joint action $a$) is the one maximizing: $\sum_{i\epsilon L} r_i + \gamma . v_i $, where $L$ is the set of all Lanes, $r_i$ is the predicted reward  for the transition $\mathcal{T}(s,a) \rightarrow s'$, and $v_i$ is the estimated (via TS) long term value for state $s'$ (i.e. the state reached after taking action $a$).

\subsubsection{Improving priors}
To train $\phi$, local actions are extracted from the best coordinated set and become training targets. As training progresses, local priors increasingly favor the sampling of local actions belonging to the best joint action set, improving TS estimates, which in turn improve priors in a positive feedback loop.

Predicting local actions belonging to the best coordinated sets based on neighborhing information (i.e. using representations obtained from the GCN) makes our model-based approach scalable as it can then be used with an arbitrary planning budget. In the extreme case of null budget (i.e. no TS), actions can still be selected by acting greedily with respect to these learned "coordinated priors".  In other words, we can simply sample a candidate set using the trained local priors without running a TS.

%To sample a promising candidate set of coordinated actions at any step in the TS: 1) Using the GCN, we propagate latent lane representations (obtained via our dynamics model) to obtain intersection level nodes corresponding to all legal actions (i.e. phase nodes), 2) we estimate "coordinated priors" independently using these intersection level representations, 3) we sample priors independently for all intersections to obtain a candidate coordinated set of actions (i.e. a candidate action in the TS). This trick makes our model-based approach scalable as it can then be used with an arbitrary planning budget. In the extreme case of null budget (i.e. no TS), actions can still be selected by acting greedily with respect to the learned "coordinated priors." 

\subsection{Architecture}
\label{sec:architecture}
Road-network objects are modeled as nodes in the GCN. The 4 types of nodes are: vehicle, lane, connection\footnote{A connection exists between two lanes if one can lead to the other given a state of connectivity.} and phase.\footnote{A phase defines a state (green/yellow/red) for all connections at a given intersection.} The main difference between the GCN architectures of IG-RL\cite{devailly2021ig} and MuJAM is that the latter uses one node per phase whereas the former uses a single node to represent the controller.
The types of edges\footnote{\label{note_edges}Every type of edge uses its own set of parameters.} are: 
\begin{itemize}
    \item Edge linking the node of a vehicle to the node of a lane it is currently on.
    \item Edge linking the node of a lane to the node of another inbound lane
    \item Edge linking the node of a lane to the node of another outbound lane
    \item Edge linking the node of a connection to the node of its inbound lane
    \item Edge linking the node of a connection to the node of its outbound lane
    \item Edge linking the node of a connection node to the node of a phase
\end{itemize}

\subsubsection{Graph Features}
\label{sec:graph_features}
\begin{table}[h!]
\centering
\begin{threeparttable}
\caption{Node features}
\label{table:nodes_features}
\begin{tabular}{cl}
\toprule
\textbf{Node type}  & \textbf{Features}                                                                                                    \\ \midrule
{Vehicle (V)}      & \textit{Current speed}, \textit{Position on lane}                                                                                       \\ \midrule
{Lane (L)}          & \textit{Length}                                                                     \\ \midrule
{Connection* (C)}    & \begin{tabular}[l]{@{}l@{}}\textit{Constraints type}, \textit{Time since last switch}, \textit{Is open},\\\textit{Is yellow}, \textit{Has priority}, \textit{Next switch open},\\\textit{Number of switches to open},
\textit{Next opening has priority}\end{tabular} \\ \midrule

{Phase (P)}    & \begin{tabular}[l]{@{}l@{}}\textit{None}\end{tabular}                                                                                             \\ \bottomrule
\end{tabular}
\end{threeparttable}
\end{table}

\begin{table}[h!]
\centering
\begin{threeparttable}
\caption{Edge features}
\label{table:edges_features}
\begin{tabular}{cl}
\toprule
\textbf{Edge type}  & \textbf{Features}                                                                                                    \\ \midrule
{V to L}      & \textit{None}\\ \midrule
{L to L}          & \textit{Is inbound} + all C features\\ \midrule
{L to C}          & \textit{Is inbound} + all C features\\ \midrule
{C to P}          & \textit{Opens connection}, \textit{Is legal}\\ \bottomrule
\end{tabular}
\end{threeparttable}
\end{table}

Node and edge features are summarized in Tables~\ref{table:nodes_features}, \ref{table:edges_features}. \textit{Current speed} represents the normalized current speed of a vehicle (between 0 and 1: speed in km/h divided by the maximum allowed speed of 50 km/h), \textit{Position on lane} represents the relative location of a vehicle on a given lane (between 0 and 1, expressed as a proportion of the lane). \textit{Length} is the length of a lane in meters, \textit{Is open} indicates if a given connection is currently opened (i.e. green), \textit{Is yellow} indicates if a given connection is currently yellow. \textit{Has priority} represents, when a connection is opened, if it has priority (i.e. if vehicles following a connection have priority over vehicles following alternative connections at an intersection). \textit{Next switch open}\footnote{\textit{Next switch open, Number of switches to open \& Next opening has priority are set to -1 when using acyclic constraints}} indicates if the connection will be opened on the next phase of the cycle. \textit{Number of switches to open}\footnotemark[\value{footnote}] indicates, in the case of cyclic constraints, in how many switches the connection will be opened (i.e. green). \textit{Next opening has priority}\footnotemark[\value{footnote}] indicates if the next opening of the connection has priority. \textit{Time since last switch} is the number of seconds since the last change in connectivity (switch) at the corresponding intersection. \textit{Is inbound} represents whether an edge (in the GCN) follows the direction of traffic (i.e. propagates information in the direction of traffic) or follows the opposite direction of traffic (i.e. propagates information in the opposite direction of traffic). \textit{Opens connection} indicates, for a connection-phase pair, if the connection is opened (i.e. green) under the corresponding phase. \textit{Is legal} indicates if a given phase can legally be selected at the current time step.

\subsubsection{Computational Graph \& Propagation rules}
\textbf{Initial Representation}:
The \textit{initial representation} for the road network aims at obtaining, for all lanes, a representation of its current surroundings (connectivity and traffic). To obtain this representation, messages are first propagated once along V-to-L edges to obtain a representation of demand on all lanes. Then, in order to contextualize this representation, messages are propagated $K$ (a hyperparameter) times along L-to-L edges. 

\textbf{Dynamics}: Based on a representation $s_t$ of the road network, and given a joint action $a_t$, to obtain the representation of the next time step (as part of a simulated trajectory), we first manually update all C features (on both nodes and edges)~(see Tab. \ref{table:nodes_features}, \ref{table:edges_features}). This informs the model of the evolution of connectivity dynamics (i.e. the immediate impact of $a_t$). Then, we propagate messages alongside L-to-L edges $K'$ (a hyperparameter) times. This last step aims at modeling traffic dynamics between lanes based on the previous representation $s_t$ and updated connectivity features. 

\textbf{Value-related-metrics computation}:
For all lanes, reward ($r$) and value ($v$) estimates for a given time step are obtained by feeding lane's representation (i.e. L nodes embeddings) to 2 respective MLPs.%

\textbf{Prior computation}:
Given a representation of the road network (i.e.  contextualized representations for all lanes), messages are propagated once along L-to-C edges, and then once along C-to-P edges. For all intersections, the obtained representations of legal phases nodes (i.e. nodes embeddings of phases which are legally selectable at the corresponding time step) are independently fed to a MLP ($\phi$) and then normalized (per intersection) using a softmax to obtain probabilities (i.e. the predicted local action distribution, used as prior during planning).

\begin{figure}[htp]
\centering
\includegraphics[width=8cm]{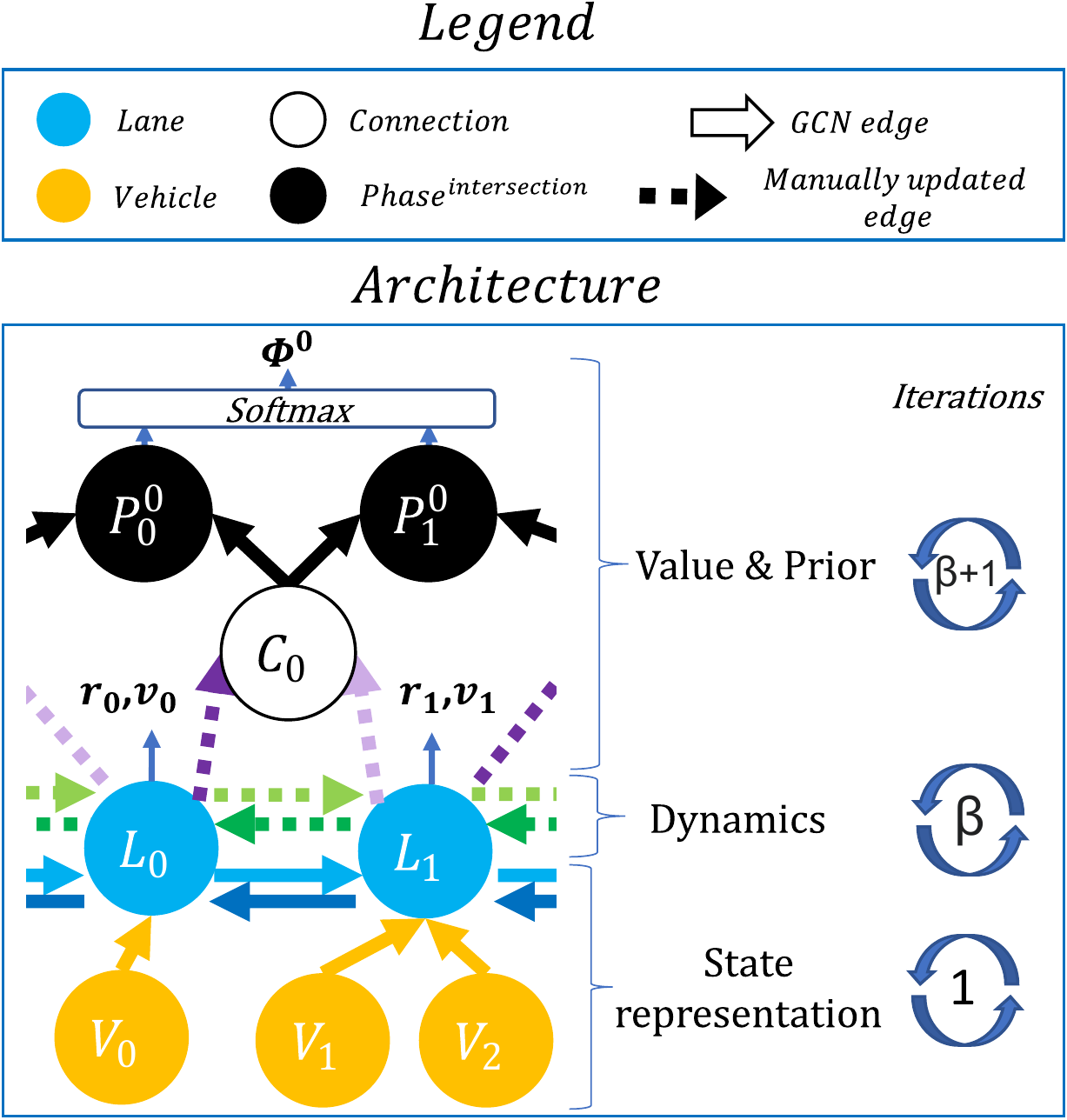}
\caption{Model: For simplicity, we illustrate the computational graph under a cyclic policy. First, we obtain the current state representation of all lanes by aggregating vehicle level information and propagating representations along connected lanes. Then, we unroll our dynamics model $\beta$ times by first manually updating connectivity-related features (i.e. \emph{connectivity dynamics}) in the GCN, and then propagating updated representations (i.e. modeling \emph{traffic dynamics}) along the updated edges. For both the initial representation and simulated steps: 1) value related metrics ($r_i$,$v_i$) are obtained using MLPs on lanes' representations, and these values are then summed to obtain network-level-estimates 2) local priors are obtained using the second part of the GCN which first computes connections representations from their respective inbound and outbound lanes' representations, and then computes phases representations from the representations of connections they control. For all intersections, final representations of all legally selectable phases (current phase and next phase in the particular cyclic setting) are then fed to a final MLP and all outputs from a given intersection are normalized using a softmax to get the final priors $\Phi$ (i.e. probabilities of selecting each phase).}
\label{fig:MuJAM}
\end{figure}

Fig.~\ref{fig:MuJAM} illustrates and describres a slice of the computational graph (i.e. how MuJAM models dynamics and predicts value-related quantities and coordinated priors).
    
\subsection{Planning with a tree search}
At any given time step, in order to select a joint action, TS can be performed according to a predefined search budget $\beta$ (a hyperparameter). If $\beta=0$, we sample an action per intersection using the local priors (distributions) obtained from the representation of the current state (i.e. original node). The combination of these local actions constitutes the joint action. Alternatively if $\beta>0$, $\beta$ trajectories will be simulated as part of the TS (trajectories must not exceed the maximum search depth $\delta$). 

\subsubsection{Growing the tree}
Inspired by the approach proposed in ~\cite{hubert2021learning} we perform guided sampling of the action space. In our coordinated setting, we use local priors, trained on coordinated targets and combine locally sampled actions to form joint actions. As the number of possible actions 1) varies dramatically across time steps (because of constraints) and across road networks 2) explodes exponentially with the number of intersections, we propose the following progressive widening strategy for the TS: If $samples~<~((actions/C1)^{C2}).(visits^{C3})$ (where $samples$ is the number of sampled actions at the current node, $actions$ is the number of selectable actions at the current node, $visits$ is the number of times the current node has been visited in the current TS, and $C1,C2,C3$ are hyperparameters), then a new joint action is sampled using all local priors computed from the current node representation. This new progressive widening strategy enables controlling the rate at which alternative trajectories are evaluated at a given node based on the size of the action space \emph{at that node}. For instance, this enables ensuring that the width of the search increases (i.e. more actions are sampled) with the action space's size, \emph{ceteris paribus}.

\subsubsection{Exploring the tree}
During the simulation of a trajectory, at a given node, an action is selected (i.e. a child node is selected) using the following probabilistic upper confidence tree (PUCT) bound~\cite{silver2016mastering}: $\left[ argmax_aQ(s,a)~+~c(s).\Phi(s, a)\dfrac{\sqrt{P_{visits}}}{1+Ch_{visits}}\right]$, with $c(s)$~=~$log((P_{visits}+C_{base}~+~1)/C_{base})+C_{init}$ where $P_{visits}$ is the number of times the parent/current node was visited, $Ch_{visits}$ is the number of times the child node (i.e. corresponding action) was visited, $\Phi$(s, a) is the joint prior (i.e. the probability of sampling the given joint action obtained by the product of local probabilities), and $C_{base},C_{init}$ are hyperparameters. Additionally, as done in \cite{schrittwieser2020mastering}, we add Dirichlet noise to local priors of the source node to favor exploration at the root of the TS. 

%$argmax_aQ(s,a) + c(s).\pi(s, a)\dfrac{\sqrt{\sum_bN(s, b)}}{1+N(s,a)}$, with $c(s) = log((\sum_bN(s, b)+C_{base}+1)/C_{base}) + C_{init}$
\subsubsection{Backpropagating values}
When simulating a new transition (i.e. exploring a new node in the TS), the estimated value of the trajectory is backpropagated along the TS. However, unlike in Monte Carlo TS (MCTS) where the estimated value of starting from a given node is based on the average of the estimated values of trajectories simulated from that node, in our experiments, we define it as the value of the best simulated trajectory, as it improved learning and convergence speeds significantly in our experiments. 
    
\subsection{Training}
MuJAM is trained end-to-end by backpropagation of prediction errors' (on $r$,$v$, and $\Phi$) using observed sequences of transitions and corresponding TS results. 

\subsubsection{Improving data efficiency}
Targets for $v$ and $\Phi$ are built using the result of TS when interacting with the environment. However, as proposed in \cite{schrittwieser2020mastering}, we continuously reanalyze episodes in the replay buffer to refresh TS estimates using updated parameters. This is done to improve sample efficiency and training speed. Furthermore, to enforce exploration in the environment during training, instead of using independent noise at each time step as done in MuZero, we use noisy networks~\cite{fortunato2017noisy} for the prior. More specifically, the weights composing the layers of the MLP used for prior computation are defined as parameterized Gaussian distributions. These weights are periodically re-sampled from these distributions. This adaptation of MBRL enables coherent exploration as the behavior of consecutive time steps is nonlinearly correlated. In consistence with~\cite{devailly2021ig}, coherent exploration improved training speed and asymptotic performance in our experiments.

\section{Experiments}
\label{sec:Experiments}

In our experiments, we use a zero-shot transfer setting proposed by~\cite{devailly2021ig} and inspired by other transfer settings in RL~\cite{oh2017zero}~\cite{higgins2017darla}.

We compare the performence of instantiations of MuJAM and several baselines on both small synthetic road networks and the road network of Manhattan.
The following experiments demonstrate that: 
\begin{itemize}
\item MuJAM enables both higher asymptotic performance and higher data efficiency
\item MuJAM enables both the use of acyclic constraints (which typically enable better performance than cyclic constraints) and the learning of policies that are agnostic to these constraints (i.e. perform well in both settings). 
\item Joint action modeling contributes to the performance of MuJAM.
\item MuJAM distributes delay between trips in a more equitable fashion. 
\end{itemize}

\subsection{General Setup}
\subsubsection{Network generation}
Road networks are randomly generated (i.e. random connectivity and structure) using SUMO~\cite{SUMO2018}. The generated networks typically include between 2 and 6 intersections. The length of edges is between 100 and 200 meters, and between 1 and 4 lanes compose a given road.
\subsubsection{Traffic generation}
Traffic generation (i.e. the generation of trips) follows asymetric trajectory distributions (probabilities for a trip to start and finish on any given lane in the road network) which are resampled every 2 minutes to ensure non-stationarity.% The \emph{regime} refers to average frequency at which vehicles are added to the network. 

\begin{table}[h!]
\centering
\begin{threeparttable}
\caption{Hyperparameters}
\label{table:hyperparameters}
\begin{tabular}{ll|ll}
$learning\ rate$ & 1e-3  & $C_{init}$ & 1.25  \\
$\omega$           & 1e4   & $K$        & 3     \\
$batch\ size$    & 16    & $K'$       & 3     \\
$C1$            & 5     & $\gamma$   & 0.997 \\
$C2$            & 0.5   & $\delta$   & 1     \\
$C3$            & 0.5   & $\beta$    & 50    \\
$C_{base}$      & 19652 & $optimizer$           & Adam  
\end{tabular}
\end{threeparttable}
\end{table}

\subsubsection{Evaluation}
We use the same networks and trips to evaluate all methods. Performance is evaluated using the instantaneous delay, defined as: $d_t=\sum_{v\epsilon V}(s_{v}^{*}-s_v)/s_{v}^{*}$, where $s_{v}^{*}=min(s_{v^*},s_l)$, $V$ is the set, at time step t, of all vehicles in the road network, $s_{v^*}$ is the maximum speed of a vehicle, $s_l$ is the maximum speed a vehicle can legally reach considering the lane it currently is on, $s_v$ is the vehicle speed at time step t.

\subsubsection{Robustness}
\label{robustness}
We run all experiments 5 times to ensure the robustness of our conclusions as random seeds influence network and traffic generation, initial parameters' values for learnable approaches, and exploratory noise during training.

\subsection{Baselines}
\subsubsection{Fixed Time}
The Fixed Time baseline follows SUMO default cyclic programs which is defined based on the structure of a given intersection

\subsubsection{Max-moving-car heuristic (Greedy)}
This dynamic baseline aims to enable the movement of a maximum number of vehicles at any given time. To do so, it switches to a next phase as soon as there are more immobilized vehicles than moving vehicles in lanes which are inbound to the intersection. Otherwise, it prolongs the current phase.

\subsubsection{IG-RL}
This is the only learnt baseline for which zero-shot transfer to new road architectures and traffic distributions is achievable. It consists in a GCN which gathers demand at the vehicle level and is trained via RL using noisy parameters for exploration~\cite{devailly2021ig}.

\subsubsection{MuJAM}
For MuJAM, 50\% of intersections used during training are under cyclic constraints and 50\% are under acyclic constraints to enforce generalization to these constraints. For MuJAM-C and MuJAM-A, all intersections used in both training and test are under the corresponding constraints (cyclic and acyclic, respectively). The suffix \emph{NNL} (no noisy layers) indicates that exploration in the environment was performed as done in the original MuZero formulation (i.e. exploratory behavior is independent between 2 consecutive time steps), instead of using noisy layers (which ensure coherence in exploratory behavior). The suffix \emph{NR} (no reanalyze) indicates that we use the default (more mainstream) instantiation of MuZero~\cite{schrittwieser2020mastering} which does not periodically reanalyzes transitions to refresh training targets with updated parameters. Finally, for MuIM (independent modeling), a TS is performed independently for each intersections instead of performing a joint TS for the entire road-network. (i.e. joint-action-modeling is removed). 

\subsubsection{Training and hyperparameters}
Training episodes last 10 minutes (simulation time). During training, performance is continuously evaluated on a separate set of 10 road networks. If a method does not reach a higher average reward on this set for $\omega$ steps, early stopping is enforced (i.e. training ends). For IG-RL, hyperparameters are the same as those used in the original paper~\cite{devailly2021ig}. 
We now refer to all methods based on MuZero as \emph{Mu methods}. For \emph{Mu methods}, hyperparameters are listed in Tables~\ref{table:hyperparameters}. and are either chosen according to other works or based on computational constraints.

\subsection{Experiment 1: Inductive Learning \& Zero-shot transfer}
\label{section:experiment1}
%\subsubsection{Simulation}
Traffic is generated for the first 10 minutes (simulation time). On average, a vehicle is introduced every 4 seconds in a given road-network. Episodes are terminated as soon as all trips have been completed. As all methods are evaluated using the same set of trips, they are paired together to study differences in delays. We report paired t-tests and distributions of differences in delays, when compared to the best performing method (MuJAM with acyclic constraints), in Fig.~\ref{fig:trips_delays} and Fig.~\ref{fig:delays_differences}, respectively.%(see Fig.~\ref{fig:delays_differences}). 

%\subsubsection{Results}
%\hfill\\
%\textbf{Model-based inductive learning}:
\subsubsection{Model-based inductive learning}
\emph{Mu methods} enable lower trips delays (lower means, medians and quartiles) compared to all baselines, as shown in Fig.~\ref{fig:trips_delays}. These methods also lead to higher asymptotic performance (i.e. lower average total delay after training is completed) (see Fig.~\ref{fig:total_delay}). With the only notable exception of MuJAM-NR-C, which is discussed later in this subsection, \emph{Mu methods} offer higher data efficiency as these methods start outperforming all baselines early in training. This demonstrates that recent advances and successes in model-based RL on games (e.g. Chess, Go, Atari games) can also be leveraged in the challenge of ATSC.

%\textbf{Constraint Agnosticism}:
\subsubsection{Constraint Agnosticism}
\begin{figure}[htp]
\centering
\includegraphics[width=9cm]{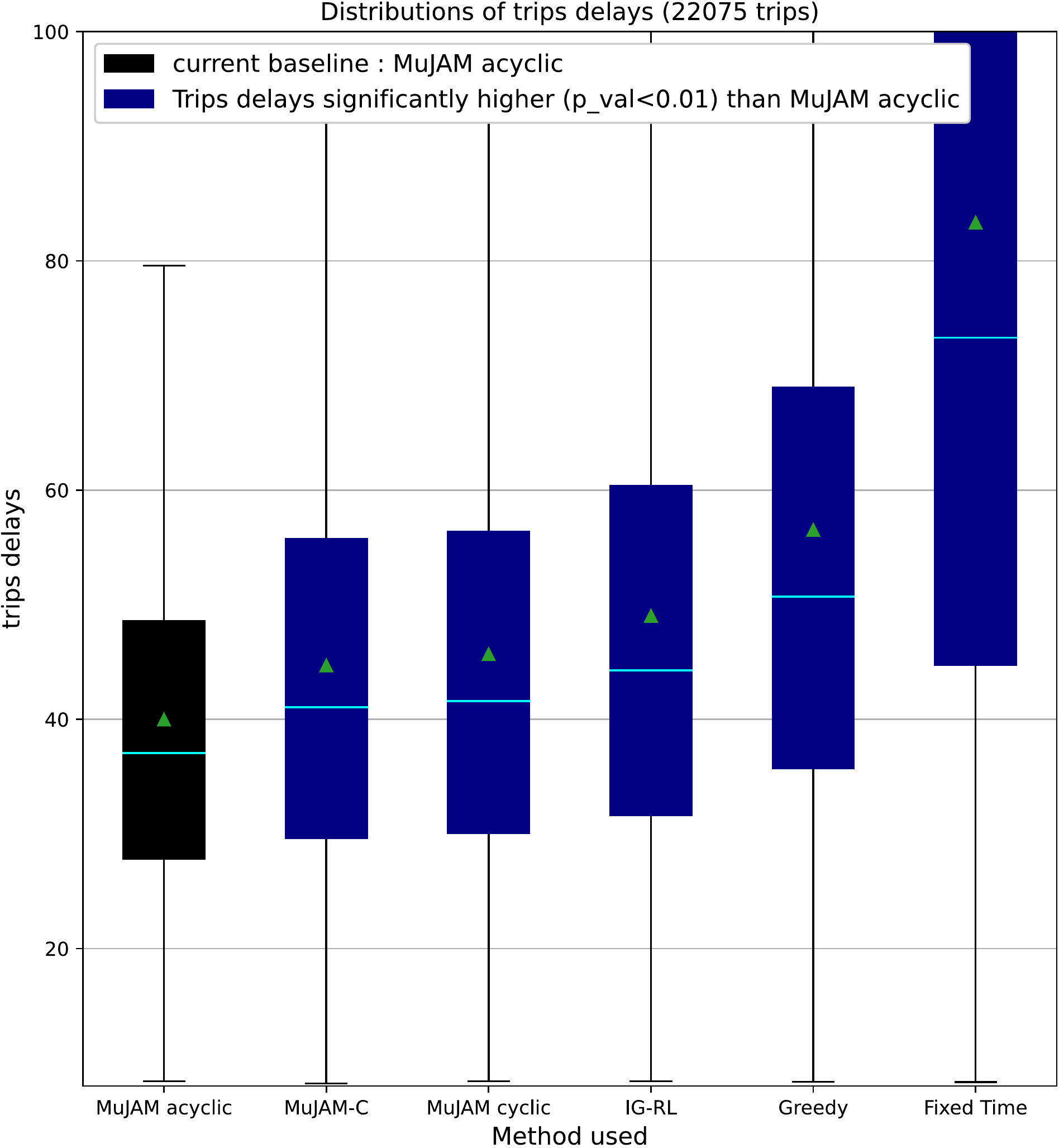}
\caption{Trips Delays | This figure displays the distributions of total delays experienced per trip.}
\label{fig:trips_delays}
\end{figure}

\begin{figure}[htp]
\centering
\includegraphics[width=9cm]{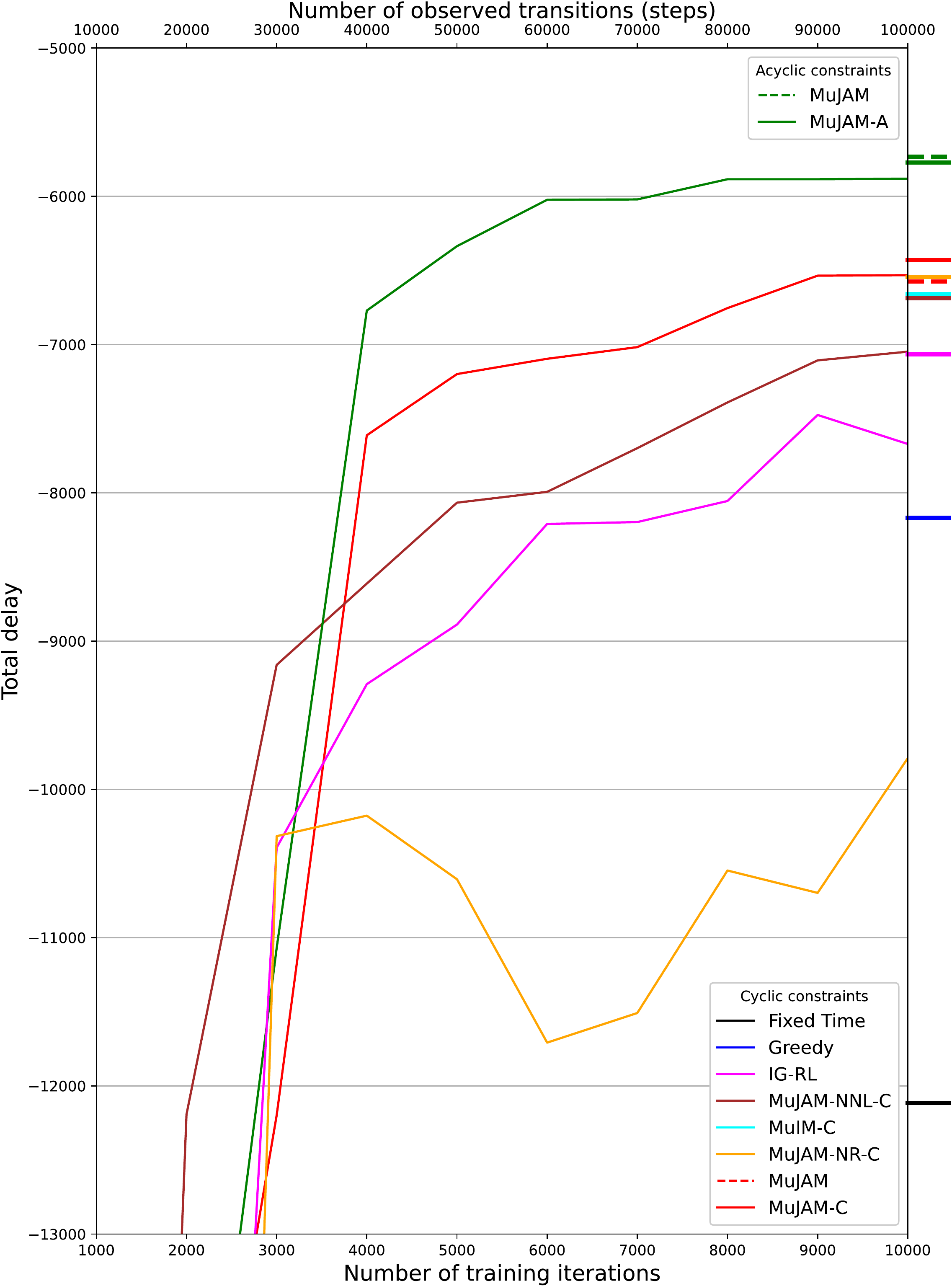}
\caption{Average total delay | This figure displays the average of the total delay (summed over all trips and all time steps) in a given episode (i.e. for a given road-network). Asymptotic performance (after training is completed) is represented as a thick line at the right extremity of the figure. For some methods (for which we specifically want to compare data efficiency) the ratio of training steps to observed transitions (i.e. interactions with the environment) is fixed to 0.1 for the first 10k training steps. Every 1k training steps in this interval, average performance of  is measured and reported.}
\label{fig:total_delay}
\end{figure}

%\subsection{Constraint Agnosticism}
First, we observe that methods under acyclic constraints outperform methods under cyclic constraints as 1) the distribution of trips delays (means, medians and quartiles) is the lowest for MuJAM with acyclic constraints (see Fig.\ref{fig:trips_delays}) and 2) both data efficiency and asymptotic performance are better for MuJAM-A than MuJAM-C (see Fig.\ref{fig:total_delay}). Although this result is not surprising as the acyclic setting is less constraining than the cyclic setting, it it the first time that an acyclic approach is transferable (with zero-shot transfer) to new road-network and traffic distributions. 
Furthermore, the evaluation of our hybrid approach, MuJAM, under both cyclic and acyclic constraints yields similar asymptotic performance compared to corresponding specialists formulations (MuJAM-A and MuJAM-C). As show in Fig.\ref{fig:total_delay}, it is slightly higher for acyclic constraints, and slightly lower for cyclic constraints. This demonstrates the ability of MuJAM to generalize to constraints and the viability of training a single agnostic method to tackle both a variety of road-network architectures, traffic distributions, and behavioral constraints simultaneously.

\begin{figure}[htp]
\centering
\includegraphics[width=9cm]{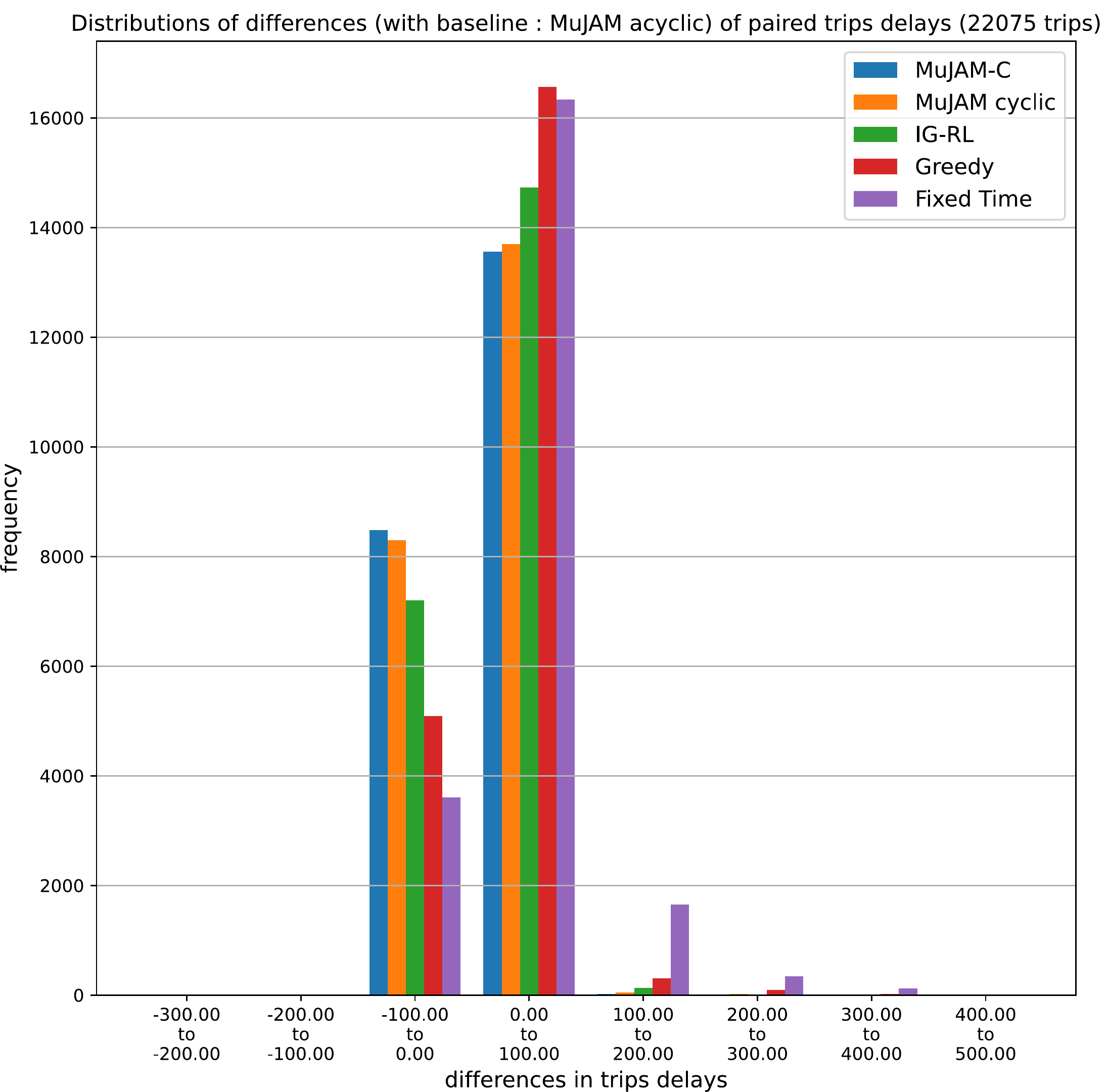}
\caption{Distributions of delays' differences | This figure displays the distributions of the differences of trips delays when compared to ~MuJAM~acyclic}
\label{fig:delays_differences}
\end{figure}

%\subsection{Ablation Studies}
%2 figures:

%1 (boxplots) with all methods: 
%IG-RL, MuJAM-C, MuIM-C, MuJAM-C(no Noisy), MuJAM-C(no Reanalyze)
%\begin{figure}[htp]
%\centering
%\includegraphics[width=8cm]{Ablation_boxplots.png}
%\caption{}
%\label{fig:MuJAM}
%\end{figure}

%and 1 (sample eff) with IG-RL, MuJAM-C, MuJAM-C(no Noisy), MuJAM-C(no Reanalyze)

%\textbf{Coordination}:
\subsubsection{Coordination}
%\subsubsection{Coordination}
The ablation of coordination ability (i.e. MuIM-C) leads to a lower asymptotic performance when compared to MuJAM-C (see Fig.~\ref{fig:total_delay}). In fact this ablation makes MuIM-C one of the worst performing \emph{Mu methods}. This demonstrates that coordination ability (jointly maximizing performance for the entire road-network) can improve performance compared to locally-greedy approaches (maximizing reward per intersection).

%\textbf{Data Efficiency}:
\subsubsection{Data Efficiency}
%\subsubsection{Data Efficiency}
Removing coherence in exploratory behavior (i.e. MuJAM-NNL-C) leads to both a slower increase in performance during training and a lower asymptotic performance when compared to MuJAM-C (see Fig.~\ref{fig:total_delay}). This ablation makes MuJAM-NNL-C the worst performing \emph{Mu method}. This demonstrates that coherence in exploration is key to both data efficiency and performance in RL-ATSC, confirming what was reported in~\cite{devailly2021ig}.
Removing the ability to refresh training targets with updated parameters (i.e. MuJAM-NR-C) negatively impacts asymptotic performance (see Fig.~\ref{fig:total_delay}), although this method still remains the second best performing \emph{Mu method} under cyclic constraints. However, this ablation impairs data efficiency the most as performance even lags behind IG-RL during the first 10k training iterations.

%\textbf{Performance Vs. Fairness}:'
\subsubsection{Performance Vs. Fairness}
ATSC distributes delay among trips in a road-network as connectivity at any intersection does not enable all vehicles to travel at full speed (i.e. some connections are always closed). An improvement in global performance should not be at the expense of equity in this distribution (i.e. it is socially unacceptable to disproportionately delay some trips for the sake of the average performance). As shown in Fig.~\ref{fig:delays_differences}, in addition to accelerating most of the trips (diminishing their delays), the best performing method (MuJAM with acyclic constraints) also happens to distribute delay in a more equitable fashion. Indeed, trips which are advantaged by this method can be drastically delayed by substituting this method with other methods, but trips which are disadvantaged by this method are only marginally delayed.

\subsection{Experiment 2: Scaling to Manhattan}

%\subsubsection{Description}
In this experiment, we push zero-shot transfer to the large scale real-world-network of Manhattan (3,971 TSCs and 55,641 lanes) using a heavy traffic \footnote{Even though the road network is real and extracted from \href{www.openstreetmaps.org}{~openstreetmaps.org}, asymmetric traffic distributions used in evaluation are not based on observations of real-world distributions in Manhattan.} (tens of thousands of vehicles with asymmetric traffic distributions) to evaluate the extent of generalizability and scalability of approaches studied in ~\ref{section:experiment1}. The combination of new network structure (including some intersections with complex patterns), new (heavier) traffic distributions, and scaling to a network much larger than anything experienced in training is expected to be a challenging task.

Zero-shot transfer is what enables running this large scale experiment with learned methods as training on such a large network (particularly if using a different set of parameter per intersection as typically done in MARL-ATSC) would involve prohibitive computational costs. 

In this experiment, traffic is generated for 30 minutes (warm-start) using the Fixed Time policy, so that the traffic is already dense\footnote{At this point, 18,000 vehicles have been inserted on average.} before evaluation starts for all approaches. At this point an episode lasts one hour (density continues to increase continuously during this time). As not all trips are completed at the end of an episode, we cannot pair them for comparison without introducing bias. We therefore report aggregated metrics. 

As the cost of evaluation on Manhattan is still more expensive than for synthetic networks used in ~\ref{section:experiment1}, we only evaluate methods on 6 instances of randomly generated traffic per run (for a total of 30 instances as we repeat all experiments 5 times as described in \ref{robustness}).

\subsubsection{Results}
\begin{figure}[htp]
\centering
\includegraphics[width=8cm]{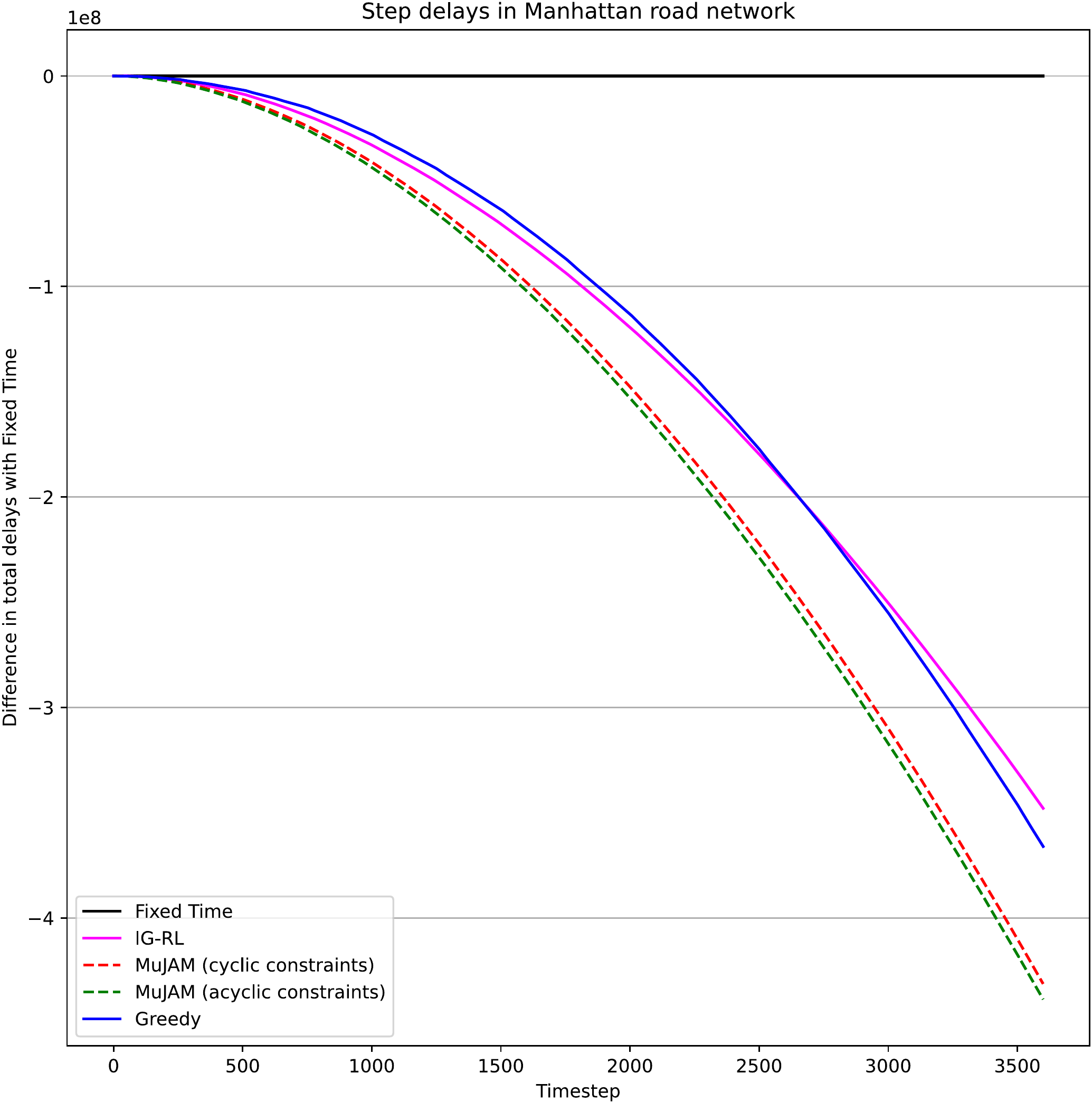}
\caption{Steps delays in Manhattan (compared to Fixed Time policy | This figure displays the cumulative difference (comparing a given method to the Fixed Time policy) in total delay experienced in the road network.)}
\label{fig:Manh}
\end{figure}

A first observation is that with MuJAM, we are able to use, for the first time, a coordinated RL-ATSC approach trained with explicit joint action modeling (not only the ability to communicate) on such a large setting. To enable this, we only use coordinated priors (see \S.\ref{coordinated_priors}) to select an action and ignore planning. In other words we use a null search budget for planning $\beta$=$0$. Under this condition, computational complexity is similar to that of IG-RL. Moreover, the same instance of MuJAM is evaluated on both cyclic and acyclic control.

Fig.~\ref{fig:Manh} displays for MuJAM, IGRL \& Greedy, the cumulative difference with Fixed Time (baseline) in total instantaneous delay per timestep for the entire duration of evaluation (one hour). All methods outperform Fixed Time. MuJAM, even under cyclic constraints, outperforms both IG-RL and the dynamic baseline. This demonstrates that in this challenging setting, this new model-based approach seems to improve generalizability and scalability. Finally, generalization to constraints is once more demonstrated as the same instantiation of MuJAM ends up outperforming all other approaches. We also note that after a certain time, IG-RL seems to underperform the dynamic baseline. We hypothethize that IG-RL suffers from a gradual shift in distribution. As density increases in the road network with time, observed distributions of traffic get further and further away from distributions used during training.

\section{Conclusion}
We introduce MuJAM, a method for zero-shot-transfer-ATSC based on MBRL and GCNs. On top of enabling representation of traffic at the finest level of granularity and transferability to new road-network-architectures and traffic distributions,  MuJAM constitutes the first bridge between learnt-cyclic and learnt-acyclic methods, which not only enables training specialist methods on either types of constraints, but also yields hybrid methods which can generalize across these constraints. It is also the first approach to enable joint action modeling at scale for RL-ATSC. MuJAM introduces MBRL developments 1) learning complex graph dynamics models in a latent space 2) improving coherence in exploration by leveraging noisy layers in MuZero, and 3) enabling multi-agent joint action modeling at scale. For RL-ATSC, this work introduces a new level of generalization in line with recent interest/development in this aspect~\cite{devailly2021ig}, which constitutes a promising way to ease experimentation in urban-mobility-planning without having to train a new model for any tweak of a behavioral constraint, which we hope contributes to real-world applicability.

%1) Cyclic and Acyclic constraints are macro-constraints. Each approach is defined using additional micro-constraints such as the  minimum time between consecutive switches or the fixed time used for phases involving yellow lights. The evaluation of simultaneous generalizability to macro-constraints and micro-constraints would provide further insights in the ability of the approaches to limit experimentation cost for urban-mobility-planning.

%\paragraph{Future work}
\subsection{Future work}
MuJAM opens a path for the following future works: 1) MuJAM offers explicit coordination. However, further investigating the advantages of said coordination and how behavior differs from independent modeling could reveal interesting patterns for RL-ATSC coordination. 2) Offline RL, which consists in training RL approaches using historical data only (i.e. no interaction with the environment), seems appealing for RL-ATSC. As MuJAM only needs to learn local-transferable-traffic-dynamics, it could prove to be less dependent on the online control of TSCs, and easier to train using offline data. Transferable-Offline RL-ATSC would enable both safe and cost-efficient training on offline data, and zero-shot transferability across networks, traffic, and constraints.

%\section{Acknowledgements}

\bibliographystyle{IEEEtran}
\bibliography{IEEEabrv,arxiv}
%\bibliography{IEEEabrv,bibliography}
%\newpage

\end{document}